\documentclass[letterpaper]{article} % DO NOT CHANGE THIS
\usepackage{aaai24}  % DO NOT CHANGE THIS
\usepackage{times}  % DO NOT CHANGE THIS
\usepackage{helvet}  % DO NOT CHANGE THIS
\usepackage{courier}  % DO NOT CHANGE THIS
\usepackage[hyphens]{url}  % DO NOT CHANGE THIS
\usepackage{graphicx} % DO NOT CHANGE THIS
\usepackage{natbib}  % DO NOT CHANGE THIS AND DO NOT ADD ANY OPTIONS TO IT
\usepackage{caption} % DO NOT CHANGE THIS AND DO NOT ADD ANY OPTIONS TO IT
\usepackage{algorithm}
\usepackage{algorithmic}

\usepackage{threeparttable}
\usepackage{tabularx}
\usepackage{tabularray}
\usepackage{booktabs}
\usepackage{verbatim}
\usepackage{amsopn}
\usepackage{amsmath}
\usepackage{multirow}
\usepackage{color}
\usepackage{colortbl}
\usepackage{graphicx}
\usepackage{makecell}
\usepackage{enumitem}
\usepackage{epsfig}
\usepackage{amssymb}
\usepackage{bbding}
\usepackage{bm}
\usepackage{xspace}
\usepackage{footnote}
\makeatletter
\DeclareRobustCommand\onedot{\futurelet\@let@token\@onedot}
\def\@onedot{\ifx\@let@token.\else.\null\fi\xspace}

\def\eg{\emph{e.g}\onedot} 
\def\ie{\emph{i.e}\onedot} 
 
\def\etc{\emph{etc}\onedot} \def\vs{\emph{vs}\onedot}
 
\def\etal{\emph{et al}\onedot}
\makeatother
\usepackage{epstopdf}

\usepackage{newfloat}
\usepackage{listings}
\DeclareCaptionStyle{ruled}{labelfont=normalfont,labelsep=colon,strut=off} % DO NOT CHANGE THIS
\floatstyle{ruled}
\newfloat{listing}{tb}{lst}{}
\floatname{listing}{Listing}
\nocopyright % -- Your paper will not be published if you use this command
\title{EulerMormer: Robust Eulerian Motion Magnification via \\Dynamic Filtering within Transformer}
\author{
Fei Wang\textsuperscript{\rm 1},
Dan Guo$^*$\textsuperscript{\rm 1,2,3},
Kun Li\textsuperscript{\rm 1},
Meng Wang$^*$\textsuperscript{\rm 1,2}
}
\affiliations{
\textsuperscript{\rm 1}School of Computer Science and Information Engineering, Hefei University of Technology (HFUT) \\
\textsuperscript{\rm 2} Institute of Artificial Intelligence, Hefei Comprehensive National Science Center \\ 
\textsuperscript{\rm 3}Anhui Zhonghuitong Technology Co., Ltd. \\
jiafei127@gmail.com, guodan@hfut.edu.cn, kunli.hfut@gmail.com, eric.mengwang@gmail.com
}

\usepackage{bibentry}
\begin{document}
\maketitle

\begin{abstract} 
Video Motion Magnification (VMM) aims to break the resolution limit of human visual perception capability and reveal the imperceptible minor motion that contains valuable information in the macroscopic domain. However, challenges arise in this task due to photon noise inevitably introduced by photographic devices and spatial inconsistency in amplification, leading to flickering artifacts in static fields and motion blur and distortion in dynamic fields in the video. Existing methods focus on explicit motion modeling without emphasizing prioritized denoising during the motion magnification process. This paper proposes a novel dynamic filtering strategy to achieve static-dynamic field adaptive denoising. Specifically, based on Eulerian theory, we separate texture and shape to extract motion representation through inter-frame shape differences, expecting to leverage these subdivided features to solve this task finely. Then, we introduce a novel dynamic filter that eliminates noise cues and preserves critical features in the motion magnification and amplification generation phases. Overall, our unified framework, EulerMormer, is a pioneering effort to first equip with Transformer in learning-based VMM. The core of the dynamic filter lies in a global dynamic sparse cross-covariance attention mechanism that explicitly removes noise while preserving vital information, coupled with a multi-scale dual-path gating mechanism that selectively regulates the dependence on different frequency features to reduce spatial attenuation and complement motion boundaries.
We demonstrate extensive experiments that EulerMormer achieves more robust video motion magnification from the Eulerian perspective, significantly outperforming state-of-the-art methods.
% The source code is available at \href{https://github.com/VUT-HFUT/EulerMormer}{https://github.com/VUT-HFUT/EulerMormer}.
The source code is available at \url{https://github.com/VUT-HFUT/EulerMormer}.
\end{abstract}

\begin{figure}[t!]
\begin{center}
\includegraphics[width=1\linewidth]{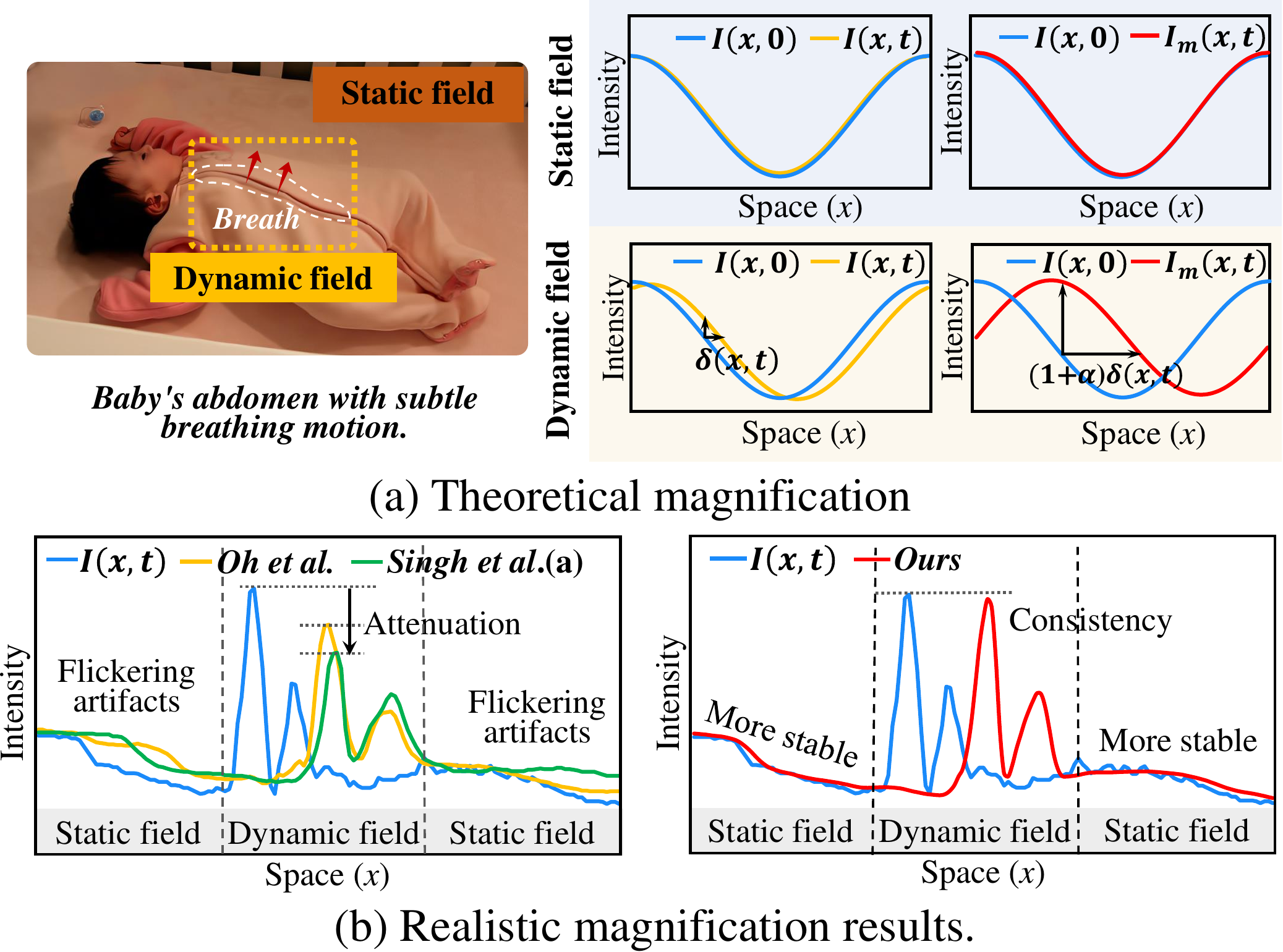}
\vspace{-2.0em}
\caption{Theoretical basis and realistic results of video motion magnification.
	Theoretically, the static field in (a) is free of position displacement, while the dynamic field should exhibit ideal position displacement to satisfy the desired motion magnification.
	However, in the real world, unavoidable photon noise and spatial inconsistency exist with flickering artifacts, intensity attenuation, \etc, as shown in (b) for the magnified results~\cite{oh2018learning,singh2023lightweight}. In contrast, our method achieves more robust magnification in both static and dynamic fields.}
\label{fig:princ}
\end{center}
\vspace{-2.5em}
\end{figure}

%%%%%%%%%%%%%%%%%%%%%%%%%%%%%%%%%%%%%%%%
\section{Introduction}
Video Motion Magnification (VMM) has garnered growing research interest due to its remarkable ability to vividly reveal subtle motions in real-world videos that are imperceptible to the human eye~\cite{rubinstein2013revealing,le2019seeing}. Existing VMM techniques behave as computer-assisted ``eyes'' that enable humans to see and grasp meaningful subtle motion in various challenging-to-perceive settings, such as micro-action  recognition~\cite{xia2020revealing,qi2020deeprhythm,mehra2022motion,Nguyen_2023_CVPR}, robotic sonography~\cite{huang2023motion}, clinical medicine~\cite{abnousi2019novel} material properties estimation~\cite{davis2015visual,davis2017visual} and modal analysis~\cite{eitner2021effect}. Specifically, VMM aims to capture and amplify the imperceptible subtle motion in the video sequence while preserving fine spatial details for realistic and accurate visualization.

However, this task faces several challenges: (1) \textbf{Photon noise}~\cite{wu2012eulerian,oh2018learning} is inevitably present in videos due to the characteristics of charge-coupled devices (CCDs) in photographic devices and signal attenuation during transmission. This noise, indistinguishable from subtle motions, can result in flickering artifacts, \etc.
(2) \textbf{Spatial inconsistency}~\cite{wadhwa2013phase,singh2023multi} measures the magnification quality, as forced motion magnification can lead to spatial frequency collapse, resulting in phenomena such as motion blur and distortion.
As the output results of recent methods~\cite{oh2018learning,singh2023lightweight} shown in Fig.~\ref{fig:princ}(b), noise amplification disrupts the static field of magnified image, and spatial intensity attenuation occurs in the dynamic field.

Inspired by the theory of fluid mechanics, early research drew from the Lagrangian and Eulerian perspectives.
Liu~\etal~\cite{liu2005motion} proposed the first Lagrangian-based approach, which involved tracking the motion trajectory of each pixel (optical flow) for motion magnification, but it was computationally expensive and sensitive to various noises.
In contrast, Eulerian approaches\cite{wu2012eulerian,wadhwa2013phase,zhang2017video,takeda2018jerk,takeda2019video,takeda2022bilateral} relied on traditional filters (such as Butterworth and Anisotropy filters) to handle the motion intensity occurring in specific regions rather than tracking every pixel throughout the video.
However, these Eulerian methods required fine-tuning numerous hyperparameters to adapt to different scenarios, which makes them impractical for real-world applications.

Developing effective VMM methods remains a compelling topic in the computer vision community.
Recently, learning-based methods~\cite{oh2018learning,singh2023lightweight,singh2023multi} utilizing different convolutional neural networks (CNN) have attained state-of-the-art (SOTA) performance.
Regardless of whether they introduce proxy model regularization or frequency domain phase to optimize their models, they essentially focus on representation learning, such as motion and phase, for generating motion-magnified videos without emphasizing prioritizing denoising.

This paper focuses primarily on addressing the denoising issue in VMM. We specially design a dynamic filter module $\mathcal{F}(\cdot)$ to address the previously mentioned photon noise and spatial inconsistency in static and dynamic fields.
Based on Eulerian theory, we disentangle texture and shape and further acquire the motion = $\triangle$shape, which is expected to leverage these subdivided features to solve this task finely.
Especially noteworthy is the to-be-magnified motion representation. 
In our framework, we utilize $\mathcal{F}(\cdot)$ to filter out noise cues from motion during the \emph{motion magnification} phase and refine the representations of texture and magnified shape during the \emph{amplification generation} phase.
Finally, compared with the limitation of existing CNN-based methods with local receptive fields, our method is equipped with Transformer architecture in the encoder and the dynamic filter $\mathcal{F}(\cdot)$, which can ensure the contextualized global relationship learning between pixels.
Overall, we provide a unified framework to filter out undesired noise cues in the representation learning of texture, shape, and motion, which results in a satisfactory magnification effect.

Our contributions can be summarized as follows:
\vspace{-0.2em}
\begin{itemize} 
\item We introduce a novel Transformer-based EVM architecture that offers better spatial consistency and fewer artifacts, motion blurs, and distortions in the magnified video.
To our knowledge, this is a pioneering effort in learning-based VMM.
\item We develop a dynamic filter implemented on a sparse attention strategy for static-dynamic field adaptive denoising and texture-shape joint refinement during the motion magnification and amplification generation phases.
\item We propose a Point-wise Magnifier, which improves the magnified representation by incorporating global nonlinear feature interactions per pixel to maintain spatial consistency and reduce flickering artifacts.
\item We collect a synthetic dataset containing magnification factors $\alpha$, Poisson noise $\lambda$ and Gaussian blur $\sigma$ to comprehensively evaluate the model's accuracy and robustness.
Extensive quantitative and qualitative experiments on synthetic and real-world datasets demonstrate our favorable performances against SOTA approaches.
\end{itemize}
\vspace{-0.5em}

% \vspace{-1 em}
\section{Related Work}
\paragraph{Traditional Methods.}
Lagrangian-based approaches~\cite{liu2005motion} pioneered this task by tracking the motion trajectory of each pixel for motion magnification, but dense optical flow computation is expensive and sensitive to noise.
Eulerian-based methods~\cite{wu2012eulerian,wadhwa2013phase,zhang2017video,takeda2018jerk,takeda2019video,takeda2022bilateral} concentrate on the specific regions where motion occurs, rather than tracking every pixel in the video.
Early Eulerian-based methods altered intensities to approximate linear magnification~\cite{wu2012eulerian} or decomposed the motion in the frequency domain~\cite{wadhwa2013phase}.
With further research, various hand-crafted filters, such as acceleration~\cite{zhang2017video}, jerk~\cite{takeda2018jerk}, anisotropy~\cite{takeda2019video}, and bilateral filters~\cite{takeda2022bilateral}, were explored.
These works rely on the predefined bandwidth for bandpass filters to amplify specific motions, but their effectiveness requires extensive hyperparameter tuning.
%%%%%%%%%%%%%%%
\begin{figure*}[!t]
\begin{center}
\includegraphics[width=1\linewidth]{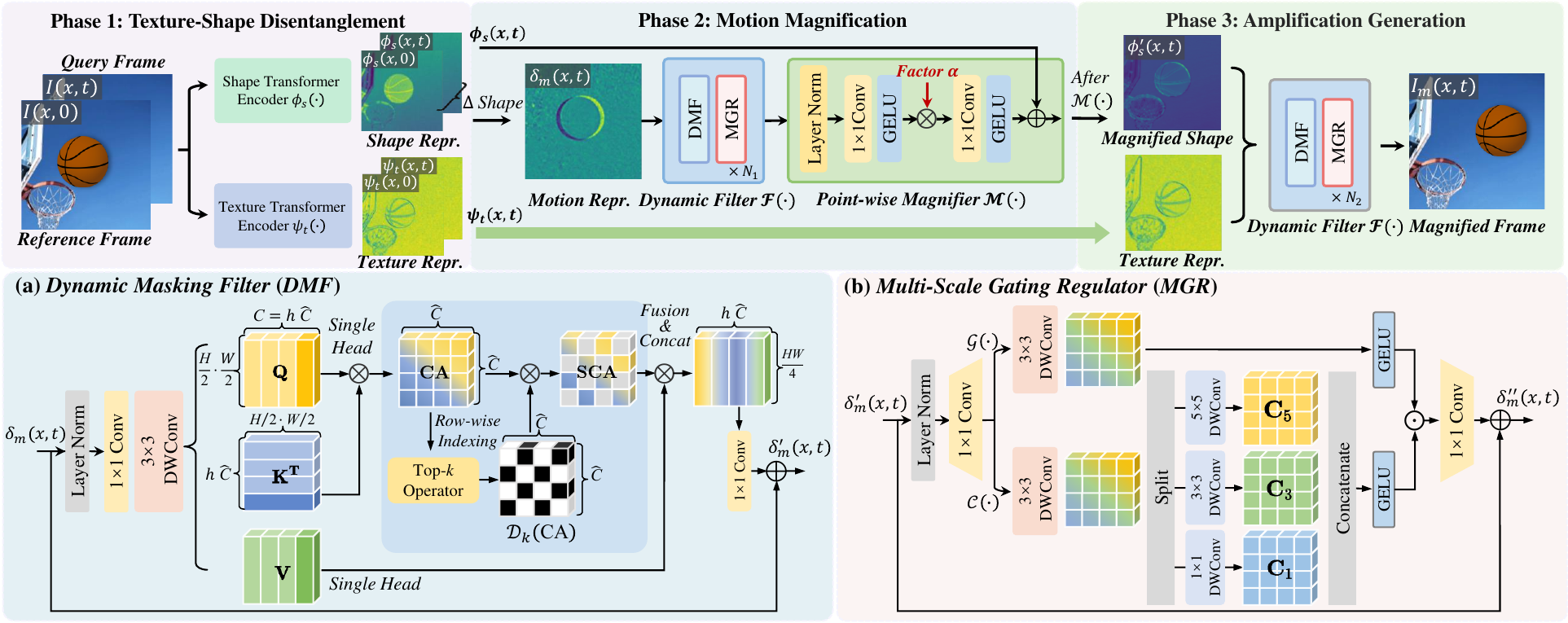}
\vspace{-1.8 em}
\caption{The overall architecture of EulerMormer for video motion magnification, which consists of three phases:
	(1) texture-shape disentanglement,
	(2) motion magnification with a dynamic filter $\mathcal{F}(\cdot)$ and a point-wise magnifier $\mathcal{M}(\cdot)$, and (3) amplification generation, which recouples and refines the original texture $\psi_t(x,t)$ and the magnified shape $\phi'_{s}(x,t)$ to generate high-quality magnified frames.
	Among them, the dynamic filter $\mathcal{F}(\cdot)$, consisting of DMF in (a) and MGR in (b), performs twice in motion magnification and amplification generation %(the refinement of the joint texture-magnified shape representation) 
	processes, which targets to achieve the static-dynamic field adaptive denoising in terms of texture, shape and motion representation learning.
}
\label{fig:arch}
\end{center}
\vspace{-2.2em}
\end{figure*}
%%%%%%%%%%%%%%%%%%%%%%%%%%%%%
\paragraph{Deep-Learning Methods.}
Learning-based approaches for the VMM have emerged but are still in their infancy, with only a handful of related works~\cite{oh2018learning,brattoli2021unsupervised,singh2023lightweight,singh2023multi}.
Oh~\etal~\cite{oh2018learning} proposed a CNN-based end-to-end architecture for the first attempt to learn the motion magnification representation, achieving comparable results to hand-crafted filters.
Recently, Singh~\etal~\cite{singh2023lightweight} proposed a lightweight CNN-based proxy model to eliminate undesired motion efficiently.
Afterwards, they~\cite{singh2023multi} also utilized CNN to model the magnification signals from frequency domain phase fluctuations to avoid artifacts and blurring in the spatial domain. Unlike the above CNN methods with local receptive fields, we introduce a novel dynamic filtering strategy into Transformer-based architecture~\cite{guo2019dadnet,li2021proposal,tang2022gloss,li2023vigt} in this study. Intrinsically, based on the Eulerian theory, our method integrates the advanced Transformer to globally model the texture, shape, and motion (inter-frame shape difference) representations, enabling static-dynamic field adaptive denoising for motion magnification.
\section{Preliminaries}
\subsection{Task Definition}
\label{sec:definition}
Let $I(x,t)$ denote the image intensity at spatial position $x$ and time $t$. With $I(x,0)$=$f(x)$ and $I(x, t)$=$f(x+\delta(x, t))$, $\delta(x,t)$ denotes a displacement function of $x$ at time $t$, the goal of VMM is to synthesize an image with respect to a magnification factor $\alpha$ as follows~\cite{wu2012eulerian}: 
\begin{equation}
\setlength{\abovedisplayskip}{3pt}
\setlength{\belowdisplayskip}{3pt}
\begin{aligned}
{I}_{m}(x,t)=f(x+(1+\alpha) \delta(x, t)).
\end{aligned}
\label{Eq:I}
\end{equation}

We can approximate $I(x, t)$ in a first-order Taylor series expansion as:
\begin{equation}
\setlength{\abovedisplayskip}{3pt}
\setlength{\belowdisplayskip}{3pt}
\begin{aligned}
I(x, t)&\approx f(x)+\delta(x, t) \frac{\partial f(x)}{\partial x},
\label{Eq:I-taylor}
\end{aligned}
\end{equation}
where $\delta(x, t) \frac{\partial f(x)}{\partial x}$ is regarded as the intensity magnitude.

Combining Eqs.~\ref{Eq:I} and \ref{Eq:I-taylor}, we have the magnification: 
\begin{equation}
\setlength{\abovedisplayskip}{3pt}
\setlength{\belowdisplayskip}{3pt}
\begin{aligned}
I_{m}(x, t)&\approx f(x)+(1+\alpha) \delta(x, t) \frac{\partial f(x)}{\partial x}.
\label{Eq:Mag}
\end{aligned}
\end{equation}

According to Eulerian learning-based VMM methods~\cite{oh2018learning,singh2023lightweight}, the motion magnification process can be disentangled into texture and shape components as shown in Eq.~\ref{Eq:LBEVM}.
In this work, our method belongs to this methodological scope. 
\begin{equation}
\setlength{\abovedisplayskip}{3pt}
\setlength{\belowdisplayskip}{3pt}
\begin{aligned}
I_{m}(x, t) \approx \underbrace{I(x,t)}_{Texture} + ~\alpha \underbrace{\delta(x, t)}_{\triangle {Shape}} \frac{\partial f(x)}{\partial x}.
\end{aligned}
\label{Eq:LBEVM}
\end{equation}

\subsection{Motivation}  
\label{sec:motivation}
As described above, video images can be modeled by two independent latent variables: texture and shape.
Texture representation exhibits invariance, while the motion generated by shape displacement for magnification deserves further investigation.
We extract subtle motion by calculating the inter-frame shape difference between two frames, \ie, motion $= \triangle$ {shape}.
Meanwhile, the amplification of subtle motion is inevitably affected by noise, as depicted in Fig.~\ref{fig:princ}, \ie, photon noise (inherent to the camera sensor)~\cite{oh2018learning} in the static field and spatial
inconsistency in the dynamic field.
To this end, we propose a dynamic filter $\boldsymbol{\mathcal{F}(\cdot)}$ in our framework designed explicitly for denoising to eliminate artifacts and distortion caused by these noises.
It is applied twice within our framework: once for denoising the motion representation and once for denoising the recoupled texture-magnified shape joint refinement, formulated as follows:
\begin{equation}
\setlength{\abovedisplayskip}{3pt}
\setlength{\belowdisplayskip}{3pt}
\begin{aligned}
I_{Ours}(x,t)=\boldsymbol{\mathcal {F}} \big[\underbrace{I(x,t)}_{Texture} + \alpha ~\boldsymbol{\mathcal {F}} \big(\underbrace{\delta(x, t)}_{\triangle {Shape}}\frac{\partial f(x)}{\partial x}\big) \big].
\end{aligned}
\end{equation}

\section{Methodology} 
\subsection{Texture-Shape Disentanglement}
\label{sec:encoder}
Given any pair of reference and query images in a video, $[I(x,0), I(x,t)$], we use a 3$\times$3 convolution layer to obtain initial feature maps $F(x,0), F(x,t)\in\mathbb{R}^{\frac{H}{2} \times \frac{W}{2} \times C}$, and further use a Texture Transformer Encoder $\psi_t(\cdot)$ and a Shape Transformer Encoder $\phi_s(\cdot)$ to obtain their texture and shape representations, \ie, $[\psi_t(x,0), \phi_s(x,0)] \in \mathbb{R}^{\frac{H}{2} \times \frac{W}{2} \times C}$, $[\psi_t(x,t), \phi_s(x,t)] \in \mathbb{R}^{\frac{H}{2} \times \frac{W}{2} \times C}$, as shown in Fig.~\ref{fig:arch}. 
Specifically, the two encoders comprise the Transformer with Multi-Dconv Head Transposed Attention (MDTA, derived from Restormer~\cite{zamir2022restormer}) and Multi-Scale Gated Regulator (MGR, see Sec.\ref{sec:filter}).
MDTA replaces multi-head self-attention (MHA) in Transformer and facilitates contextualized global interaction between pixels by incorporating depth-wise convolutions and cross-covariance attention.
This choice enables efficient pixel-grained representation learning, making it well-suited for this task.
Our MGR utilizes the multi-scale dual-path gating mechanism to selectively integrate features at different frequencies, providing satisfactory texture and shape representations.

\subsection{Motion Magnification} \label{sec:filter}
Obtaining a ``clean'' motion representation is crucial for motion magnification, as the inherent photon noise has nearly equivalent energy fields and the subtle motion change and is prone to amplify noise resulting in artifacts and distortion.
We define the motion representation by implementing a simple inter-frame shape difference, 
\ie, $\delta_{m}(x,t) = \triangle (\phi_s(x,t), \phi_s(x,0))  \in \mathbb{R}^{\frac{H}{2} \times \frac{W}{2} \times C}$. 
To manipulate the motion magnification, we describe two core components (DMF and MGR, see below) of the dynamic filter $\mathcal{F}(\cdot)$ and a point-wise magnifier $\mathcal{M}(\cdot)$ in detail below.

\subsubsection{Dynamic Masking Filter (DMF).}
We revisit multi-head self-attention on the motion $\delta_m(x,t)$. After implementing $1\times1$ convolution and $3\times3$ depth-wise convolutions, we group 
$\delta_m(x,t)$ into $h$ heads and each single-headed projection has $\mathbf{Q}, \mathbf{K}, \mathbf{V} \!\in\!\mathbb{R}^{(\frac{H}{2} \times \frac{W}{2})\times \hat{C}}$, where $\hat{C}=\frac{C}{h}$ and $h$ = 4.
On each head, we calculate a cross-covariance attention matrix $\mathbf{CA} \in \mathbb{R}^{\hat{C} \times \hat{C}}$ between $\mathbf{K}$ and $\mathbf{Q}$.
In CA, a learnable temperature $\tau$ scales inner products before calculating attention weights, enhancing training stability. 
\begin{equation}
\setlength{\abovedisplayskip}{3pt}
\setlength{\belowdisplayskip}{3pt}
\begin{aligned}
\mathbf{CA}=\tau \mathbf{K}^{\mathrm{T}}\cdot\mathbf{Q},
\end{aligned}
\end{equation}

In Fig.~\ref{fig:arch}(a), a critical design of DMF is that we take $\mathbf{CA}$ as a search space to perform dynamic sparse erasing. Our sparse strategy applies a dynamic filtering mechanism with the Top-$k$ operator~\cite{zhao2019explicit,wang2022kvt} along the channel dimension.
Specifically, we adaptively select row-wise top-$k$ contributive elements based on the channel correlation scores in $\mathbf{CA}$. Then, we utilize Eq.~\ref{Eq:mask} to generate the corresponding binary mask for position indexing, representing the relative positions of the high-contributing elements obtained in $\mathbf{CA}$.
Here, the dynamic mask $\mathcal{D}_{k} (\mathbf{CA})\in \mathbb{R}^{\hat{C} \times \hat{C}}$ is formulated as:
\begin{equation}\label{Eq:mask}
\setlength{\abovedisplayskip}{3pt}
\setlength{\belowdisplayskip}{3pt}
\begin{aligned}
\left[\mathcal{D}_k(\mathbf{CA})\right]_{i j}= \begin{cases}\mathbf{CA}_{i j} & \mathbf{CA}_{i j} \geq k_{i j} \\ 0 & \text { otherwise }\end{cases},
\end{aligned}
\end{equation}
where $k_{i j}$ represents the $k$-th row-wise maximum value in $\mathbf{CA}_{i j}$. This allows us to dynamically degenerate the dense $\mathbf{CA}$ into a sparse attention matrix $\mathbf{SCA} \in \mathbb{R}^{\hat{C} \times \hat{C}}$:
\begin{equation}
\setlength{\abovedisplayskip}{3pt}
\setlength{\belowdisplayskip}{3pt}
\begin{aligned}
\mathbf{SCA}=\operatorname{Softmax}(\mathcal{D}_{k} (\mathbf{CA})).
\end{aligned}
\end{equation} 

After the implementation of the weighted \emph{value} $\mathbf{V}$ sum with the sparse matrix $\mathbf{SCA}$, we concat all the heads' results and output the updated motion $\delta'_{m}(x,t)\in \mathbb{R}^{\frac{H}{2} \times \frac{W}{2} \times C}$.
DMF is designed to explicitly remove noise from the static-dynamic fields in $\delta_{m}(x,t)$ and preserve the desired motion, preventing distortion and artifacts caused by amplified noise.

\subsubsection{Multi-Scale Gating Regulator (MGR).}
Humans intelligently perceive visual changes across multiple scales. However, when the motion is too subtle and indistinguishable from noise, the integrity of the motion trajectory is compromised.
Based on the DMF processing noise, we propose MGR that repairs the smoothness and uncertainty of the motion contours to overcome this issue.
The MGR is a dual-path feedforward network consisting of multi-scale context branches $\mathcal{C}(\cdot)$ and dual-path gating $\mathcal{G}(\cdot)$, see Fig.~\ref{fig:arch}(b).

We normalize and map the motion $\delta'_{m}(x,t)$ to a high-dimensional space with a 1$\times$1 convolution, where the expansion factor is $\eta$ = 3. Next, after a 3$\times$3 depth-wise convolution, the motion representation is split into dual-path gates, \ie, $\mathcal{G}(\delta'_{m}(x,t)), \mathcal{C}(\delta'_{m}(x,t)) \in \mathbb{R}^{\frac{H}{2} \times \frac{W}{2} \times\frac{\eta C}{2}}$. 
For $\mathcal{C}(\cdot)$, we parallelly employ three depth-wise convolutions with the kernel sizes of $s\in\{1, 3, 5\}$ to capture the interactions at different frequencies $\mathbf{C}_1, \mathbf{C}_3, \mathbf{C}_5 \in \mathbb{R}^{\frac{H}{2} \times \frac{W}{2} \times\frac{\eta C}{6}}$.
Notably, high-frequency noise characterized by a small scale is effectively handled by the low-frequency characteristics of $\mathbf{C}_1$.
With increasing kernel sizes, $\mathbf{C}_3$ and $\mathbf{C}_5$ play a crucial role in motion contours acquisition and motion complementation.
And these different frequency features are fused before passing through a layer with a nonlinear activation function of GELU.
As for $\mathcal{G}(\cdot)$, a GELU activation function ensures nonlinear feature transformation. Finally, MGR regulates the output by Hadamard product $\odot$ with $\mathcal{G}(\cdot)$ and $\mathcal{C}(\cdot)$:
\begin{equation}
\setlength{\abovedisplayskip}{3pt}
\setlength{\belowdisplayskip}{3pt}
\begin{aligned}
\delta''_{m}(x,t)=\mathcal{G}(\delta'_{m}(x,t)) \odot \mathcal{C}(\delta'_{m}(x,t)),
\end{aligned}
\end{equation}
where the output of updated motion $\delta''_{m}(x,t)\in \mathbb{R}^{\frac{H}{2} \times \frac{W}{2} \times C}$. The combination process of DMF and MGR is defined as the dynamic filter $\mathcal{F}(\cdot)$.

\subsubsection{Point-wise Magnifier (PWM).} In this part, PWM serves as a manipulator to perform nonlinear magnification on $\delta''_{m}(x,t)=\mathcal{F}(\delta_m(x,t))$.
It adopts a simple and efficient design with two fundamental modifications to improve magnified representation learning:
(a) in order to reduce flickering artifacts, we abandon local convolutions and operate point-wise convolutions to interact with magnification across channels, thereby reducing checkerboard artifacts and being more compatible with global filtering;
(b) we use the more stable GELU activation function to provide nonlinear representation learning and avoid gradient explosion.
Therefore, the calculation process of PWM is as follows:
\begin{equation}
\setlength{\abovedisplayskip}{3pt}
\setlength{\belowdisplayskip}{3pt}
\begin{aligned}
\phi'_s(x,t)=W_p(\alpha \cdot W_p \cdot
\delta''_{m}(x,t))+\phi_s(x,t),
\end{aligned}
\end{equation}
where $\phi'_s(x,t) \in \mathbb{R}^{\frac{H}{2} \times \frac{W}{2} \times C}$ represents the amplified shape representation with the factor $\alpha$ and $W_p(\cdot)$ denotes the point-wise convolution with GELU activation layer.

\subsection{Amplification Generation}\label{sec:decoder}
We reconstruct the high-quality magnified image by recoupling the magnified shape $\phi'_s(x,t)$ with the original texture $\psi_t(x,t)$. Its challenge is avoiding high-frequency noise from $\psi_t(x,t)$ and ringing artifacts at the recoupled boundaries.
For this purpose, we recouple $\phi'_s(x,t)$ and $\psi_t(x,t)$ across the feature channels 
and adopt the same dynamic filter $\mathcal{F}(\cdot)$ to perform the texture-magnified shape joint refinement to facilitate their fusion and boundary completeness:
\begin{equation}
\begin{aligned}
I_m(x,t)=W_{up}\big (\mathcal{F}\big (\phi'_s(x,t),\psi_t(x,t)\big)\big),
\end{aligned}
\end{equation}
where $W_{up}(\cdot)$ denotes a layer that combines pixel shuffling operation~\cite{shi2016real} and a 3$\times$3 convolution to perform sub-pixel level upsampling, generating the final magnified image $I_m(x,t)$.
Methodologically, $\mathcal{F}(\cdot)$ in this section dynamically filters $\phi'_s(x,t)$ and $\psi_t(x,t)$ through interactive guidance along the channel to suppress noise while aiding in synthesizing smooth motion boundaries and clear details.

\begin{figure}[!t]
\begin{center}
\includegraphics[width=1\linewidth]{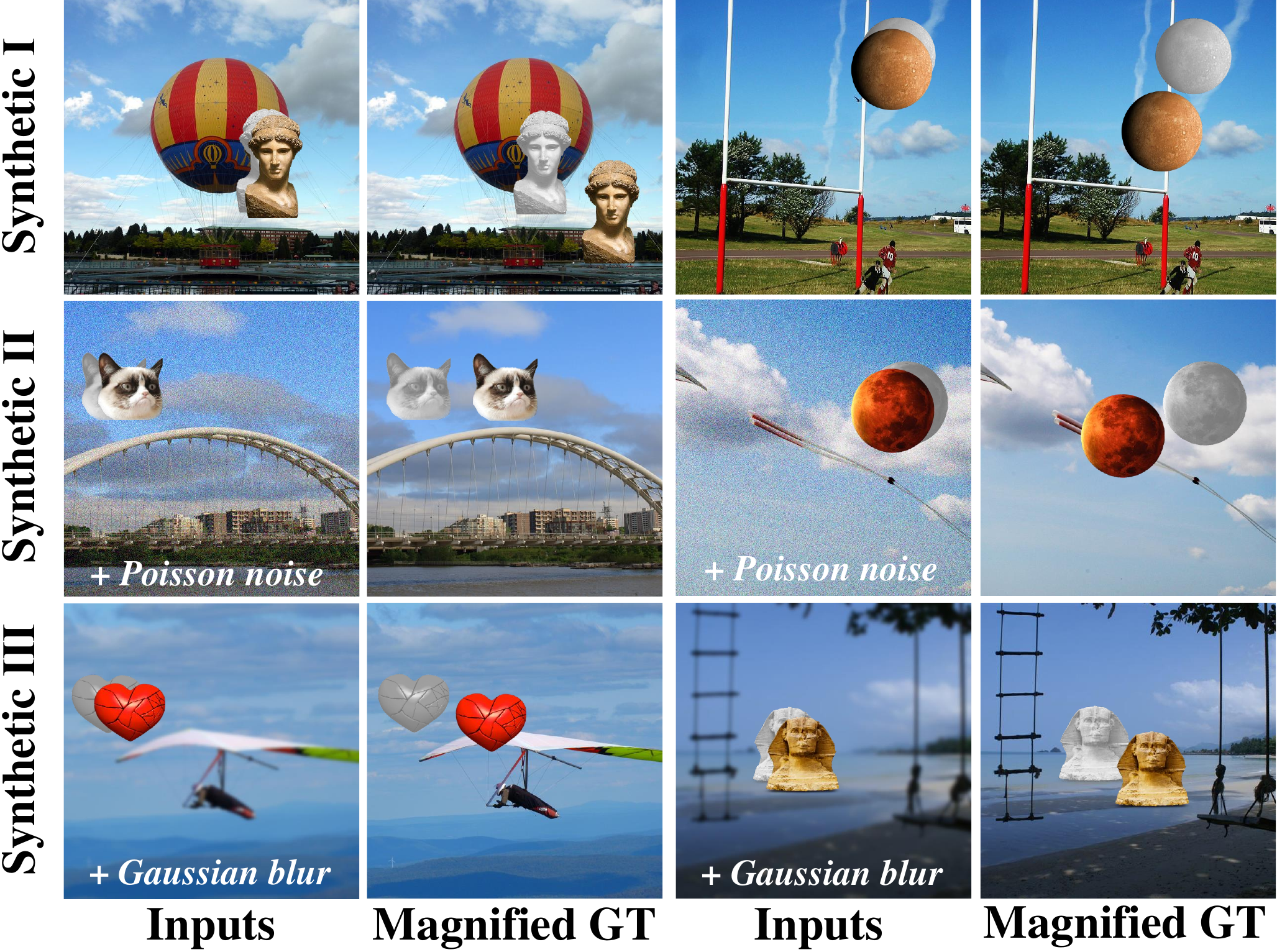}
\vspace{-1.8em}
\caption{Visualization examples of the synthetic dataset: Synthetic-\uppercase\expandafter{\romannumeral1}, Synthetic-\uppercase\expandafter{\romannumeral2} (Poisson noise) and Synthetic-\uppercase\expandafter{\romannumeral3} (Gaussian blur) datasets. To clarify the motion changes of foreground objects, we mark their reference position as grey.
}
\label{fig:syn}
\end{center}
\vspace{-2.2 em}
\end{figure}
% \vspace{-1.5em}
\subsection{Loss Optimization}\label{sec:loss} 
To optimize the proposed model, the objective function $\mathcal L$ is the weighted sum of three loss terms as follows: 
\begin{equation}
\setlength{\abovedisplayskip}{3pt}
\setlength{\belowdisplayskip}{3pt}
\begin{aligned}
\mathcal L = \mathcal{L}_{mag}+\mu_{1} \mathcal{L}_{dr} +\mu_{2} \mathcal{L}_{edge},
\end{aligned}
\end{equation}
where $\mu_{1}$, $\mu_{2}$ are hyperparameters to balance the three loss functions. 
$\mathcal{L}_{mag}$ is a basic loss term, which calculates the Charbonnier penalty~\cite{bruhn2005lucas}
between the predicted magnified image $I_{m}$ and ground-truth $I_{GT}$, formulated as follows: 
\begin{equation}
\setlength{\abovedisplayskip}{3pt}
\setlength{\belowdisplayskip}{3pt}
\begin{aligned}
\mathcal{L}_{mag}=\sqrt{\left\|I_m(x,t)-I_{GT}(x,t)\right\|^{2}+\varepsilon^{2}},
\end{aligned}
\end{equation}
where $\varepsilon$ is a constant value, being empirically set to $10^{-3}$. 
The robust Charbonnier penalty term approximates the $l_1$-loss and easily captures outliers in $I_m(x,t)$.
Besides, similar to \cite{oh2018learning,singh2023lightweight}, we use a color perturbation loss $\mathcal{L}_{dr}$ to enforce the disentangled representation learning of shape and texture as follows:
\begin{equation}
\begin{aligned}
\mathcal{L}_{dr}=\mathcal{L}(\phi_s(x,t),{\phi}^c_s(x,t)) +\mathcal{L}(\psi_t(x,t),{\psi}^c_t(x,t)).
\end{aligned}
% \vspace{-0.5 em}
\end{equation}
where $[\phi_s(x,t),\psi_s(x,t)]$ and $[{\phi}^c_t(x,t),{\psi}^c_t(x,t)]$ are respective shape and texture representations of image $I(x,t)$ and its color perturbed image $I^c(x,t)$.

Notably, in this study, we propose a new loss term $\mathcal{L}_{edge}$, namely using a Laplacian of Gaussian (LoG) edge detector $E_{LoG}$~\cite{zhang2017edge} with Charbonnier penalty, that is used to restrict the consistency between texture and amplified shape deformation as follows: 
\begin{equation}
\setlength{\abovedisplayskip}{3pt}
\setlength{\belowdisplayskip}{3pt}
\begin{footnotesize}
\begin{aligned}
	\hspace{-2mm}
	{\mathcal{L}_{edge}=\sqrt{\left\|E_{LoG}(I_m(x,t))-E_{LoG}(I_{GT}(x,t))\right\|^{2}+\varepsilon^{2}}.}
\end{aligned}
\end{footnotesize}
\end{equation}

In summary, we provide a comprehensive optimization objective for achieving the desired magnification effect, considering the factors of magnified images, texture-shape disentanglement, and recoupling consistency.

\begin{table*}[!th]
\resizebox{1.0\linewidth}{!}{
\begin{tabular}{c|c|cccc|cccc|cccc}
	\toprule
	\multirow{2}{*}{Method} & \multirow{2}{*}{Venue}
	& \multicolumn{4}{c|}{Synthetic-\uppercase\expandafter{\romannumeral1}: Magnification ($\alpha$)}
	& \multicolumn{4}{c|}{Synthetic-\uppercase\expandafter{\romannumeral2}: Poisson Noise ($\lambda$)}
	& \multicolumn{4}{c}{Synthetic-\uppercase\expandafter{\romannumeral3}: Gaussian Blur ($\sigma$)}  \\ 
	&                                 
	& RMSE$\downarrow$ & PSNR$\uparrow$ & SSIM$\uparrow$ & LPIPS$\downarrow$
	& RMSE$\downarrow$ & PSNR$\uparrow$ & SSIM$\uparrow$ & LPIPS$\downarrow$
	& RMSE$\downarrow$ & PSNR$\uparrow$ & SSIM$\uparrow$ & LPIPS$\downarrow$ \\ \midrule
	Linear%~\cite{wu2012eulerian}
	& SIGGRAPH'12 & 0.1029 & 20.21 & 0.8397 & 0.3247 & 0.1102 & 19.39 & 0.6746 & 0.2497 & 0.1347 & 17.21 & 0.5874 & 0.4666 \\
	Phase%~\cite{wadhwa2013phase}
	& SIGGRAPH'13 & 0.0978 & 21.18 & 0.8613 & 0.1428 & 0.1053 & 20.30 & 0.6941 & 0.2283 & 0.1206 & 18.87 & 0.6109 & 0.4499 \\
	Acceleration%~\cite{zhang2017video}
	& CVPR'17 & 0.0781 & 22.99 & 0.9299 & 0.1346 & 0.0854 & 22.20 & 0.7694 & 0.1922 & 0.1011 & 20.62 & 0.6508 & 0.4242 \\ 	 	 	 	 	
	Jerk-Aware%~\cite{takeda2018jerk}
	& CVPR'18 & 0.0746 & 23.61 & 0.9333 & 0.1302 & 0.0787 & 23.06 & 0.7964 & 0.1844 & 0.0951 & 20.82 & 0.6612 & 0.4156 \\ 	 	 	 	 	 
	LBVMM%~\cite{oh2018learning}
	& ECCV'18 & 0.0682 & 23.89 & 0.8748 & 0.1775 & 0.0700 & 23.65 & 0.8329 & 0.2164 & 0.0913 & 21.19 & 0.6645 & 0.4177 \\
	Anisotropy%~\cite{takeda2019video}
	& CVPR'19 & 0.0687 & 24.01 & \underline{0.9386} & 0.1260 & 0.0745 & 23.72 & 0.8230 & 0.1744 & 0.0919 & 20.93 & \underline{0.6646} & \underline{0.4121} \\
	LNVMM%~\cite{singh2023lightweight}
	& WACV'23 & 0.0662 & 24.19 & 0.8943 & 0.1544 & 0.0681 & 23.92 & 0.8497 & 0.1889 & 0.0915 & 21.16 & 0.6581 & 0.4264 \\
	MDLMM%~\cite{singh2023multi}
	& CVPR'23 & \underline{0.0615} & \underline{24.84} & 0.9173 & \underline{0.1228} & \underline{0.0637} & \underline{24.53} & \underline{0.8659} & \underline{0.1720} & \underline{0.0896} & \underline{21.34} & 0.6639 & 0.4205 \\\midrule
	\textbf{Ours}& - & \textbf{0.0594} & \textbf{25.49} & \textbf{0.9536} & \textbf{0.0535} & \textbf{0.0616} & \textbf{25.04} & \textbf{0.8706} & \textbf{0.1604} & \textbf{0.0867} & \textbf{21.89} & \textbf{0.6797} & \textbf{0.4077} \\ \bottomrule
\end{tabular}
}
\vspace{-1.0em}
\caption{Quantitative comparison of our EulerMormer and existing methods on three subsets of the synthetic dataset: evaluating magnification accuracy, noise robustness, and blur sensitivity. 
EulerMormer achieves the best performance.}
\label{tab:sys}
\vspace{-1.8em}
\end{table*}
%%%%%%%%%%%%%%%%%%%%%%%%%%%%%%%%%%%%%%%%
\begin{table}[!t]
\centering
\begin{minipage}{0.55\linewidth}
\centering
\resizebox{1\linewidth}{!}{
	\begin{tabular}{c|ccc} 
		\toprule
		Method & Static  & Dynamic & Fabric  \\\midrule
		Linear%~\cite{wu2012eulerian}    
		& 0.6288     & 0.5169   & 0.6597   \\
		Phase%~\cite{wadhwa2013phase}    
		& 0.6696    & 0.5861  & 0.7120    \\
		Acceleration%~\cite{zhang2017video}
		& 0.6748  & 0.6289  & 0.7225    \\
		Jerk-Aware%~\cite{takeda2018jerk}  
		& 0.6769   & 0.6594 & 0.7256 \\
		LBVMM%~\cite{oh2018learning}   
		& 0.6830  & 0.6409  & 0.7234 \\
		Anisotropy%~\cite{takeda2019video} 
		& \underline{0.6872}    & \underline{0.6634}   & \underline{0.7288}   \\ 
		LNVMM%~\etal~\cite{singh2023lightweight} 
		& 0.6332 & 0.6435  & 0.7195 \\ 
		MDLMM%~\etal~\cite{singh2023multi}  
		& 0.6297 & 0.6150  & 0.7134 \\ \midrule
		\textbf{Ours}   & \textbf{0.6920}   & \textbf{0.6760}  & \textbf{0.7316}  \\
		\bottomrule
	\end{tabular}
}
\vspace{-1 em}
\caption{Quantitative comparison on real-world datasets in the term of MANIQA$\uparrow$ assessment.}
\label{tab:maniqa}
\vspace{-1 em}
\end{minipage}
\hfill
\begin{minipage}{0.44\linewidth}
\centering
\includegraphics[width=\linewidth]{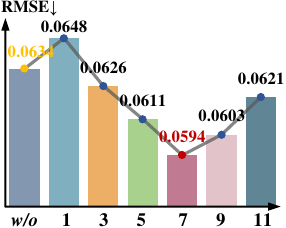}
\vspace{-1.8 em}
\captionof{figure}{Ablation results of $k$ in Top-$k$ operator on the Synthetic-I dataset. 
}
\label{fig:topk}
\vspace{-1 em}
\end{minipage}
\end{table}
%%%%%%%%%%%%%%%%%%%%%%%%%%%%%%%%%%%%
\section{Experiments}
\subsection{Experiment Setup} \label{sec:setup}
\subsubsection{Real-World Datasets.}
We experiment on three real-world benchmarks 
used in previous work:
\textbf{(a) Static dataset}~\cite{wu2012eulerian,wadhwa2013phase,oh2018learning} and \textbf{(b) Dynamic dataset}~\cite{zhang2017video,oh2018learning} contain 10 and 6 classic subtle videos in both static (slight motion, \eg baby breathing) and dynamic (\eg, strenuous motion and perspective shifts) scenarios. \textbf{(c) Fabric dataset}~\cite{davis2015visual,davis2017visual} contains 30 videos of subtle changes in fabric surface under wind excitation.

\subsubsection{Synthetic Dataset.} 
Real-world videos are rich in perceptual characteristics but lack ground truth annotations. Thus, we generate a synthetic dataset for quantitative evaluation. We select 100 objects from the public StickPNG library and 600 high-resolution background images from the DIS5K dataset~\cite{qin2022highly}.
In the data generation, we randomly place the objects onto the background images, initializing them as reference frames.
Subsequently, we synthesize query frames by randomly adjusting the objects' direction and velocity with velocities limited to the range of (0, 2] to imitate subtle motions of objects.
Therefore, we multiply the inter-frame velocities by magnification factors to synthesize the accurate ground truth for magnified motion.
We create three synthetic subsets:
\textbf{Synthetic-\uppercase\expandafter{\romannumeral1} Dataset}: Implementing random magnification factors $\alpha \in$ (0, 50];
\textbf{Synthetic-\uppercase\expandafter{\romannumeral2} Dataset}: Adding Poisson noise with the scale of random intensity levels $\lambda \in$ [3, 30]; \textbf{Synthetic-\uppercase\expandafter{\romannumeral3} Dataset}: Adding Gaussian blurs with the scale of random standard deviations $\sigma \in$ [3, 30].
In conclusion, the synthetic dataset contains 1,800 pairs of images and corresponding $\alpha$.
\textbf{We will release the data source and code on the public website.}

\subsubsection{Implementation Details.}
The focus in this field revolves around cross-dataset testing. Following the protocol~\cite{oh2018learning,singh2023lightweight,singh2023multi}, we train the model on a synthetic dataset~\cite{oh2018learning} comprising 100,000 pairs of input data sized 384 $\times$ 384 pixels.
We employ the Adam optimizer~\cite{Adam,qian2023dual} with the learning rate of 2$\times$$10^{-4}$ and the batch size of 4.
For the network hyperparameters setting, the feature channel $C$ is set to 48, and the numbers of the Texture Transformer Encoders and Shape Transformer Encoders are 2.
The dynamic filter $\mathcal{F}(\cdot)$ is configured with $N_1$ = 2 in Phase 2 and $N_2$ = 8 in Phase 3, and the Top-$k$ operator is set with $k$ = 7.
Additionally, we set the loss hyperparameters as $\mu_{1}$ = 0.1 for $\mathcal{L}_{dr}$ and $\mu_{2}$ = 0.5 for $\mathcal{L}_{edge}$.

\subsubsection{Evaluation Metrics.} 
For synthetic datasets, we employ RMSE to assess magnification error and PSNR, SSIM, and LPIPS~\cite{zhang2018unreasonable} to assess the magnification quality. 
For real-world datasets, we introduce an advanced no-reference image quality assessment metric, MANIQA~\cite{yang2022maniqa}. MANIQA is the NTIRE 2022 NR-IQA challenge winner and achieves human-comparable quality assessment and is widely applied in image distortion and video reconstruction tasks~\cite{wu2022animesr,ercan2023evreal}.

%%%%%%%%%%%%%%%%%%%%%%%%% Table3 Component
\begin{table}[th]
\begin{center}
% \resizebox{0.45\textwidth}{!}{
	\resizebox{1.0\linewidth}{!}{
		\begin{tabular}{cc|cc|cccc}
			\toprule
			\multicolumn{2}{c|}{$\mathcal{F}(\cdot)$ in Phase 2} & \multicolumn{2}{c|}{$\mathcal{F}(\cdot)$ in Phase 3} & \multirow{2}{*}{RMSE$\downarrow$} & \multirow{2}{*}{PSNR$\uparrow$} & \multirow{2}{*}{SSIM$\uparrow$} & \multirow{2}{*}{LPIPS$\downarrow$} \\
			DMF  & MGR  & DMF  & MGR  & & & & \\ \midrule
			\Checkmark   & \Checkmark & \XSolidBrush   & \XSolidBrush  &  0.0747  & 23.04 & 0.8195 & 0.2479\\
			\Checkmark   & \Checkmark & \XSolidBrush   & \Checkmark & 0.0638 &   23.38 & 0.9389 & 0.0876\\
			\Checkmark   & \Checkmark & \Checkmark   & \XSolidBrush &  0.0631  & 24.58 & 0.9437 &  0.0691 \\\midrule
			\XSolidBrush   & \XSolidBrush  & \Checkmark   & \Checkmark  &  0.0708  &  23.40  &  0.8756 &  0.1170 \\ 
			\XSolidBrush & \Checkmark & \Checkmark   & \Checkmark  &  0.0622 & 24.72  & 0.9450 &  0.0685\\
			\Checkmark   & \XSolidBrush  & \Checkmark   & \Checkmark & \underline{0.0603} &  \underline{25.34} & \underline{0.9501}  & \underline{0.0583}\\\midrule
			\Checkmark   & \Checkmark & \Checkmark   & \Checkmark & \textbf{0.0594} & \textbf{25.49} & \textbf{0.9536} & \textbf{0.0535} \\ \bottomrule
		\end{tabular}
	}
\end{center}
\vspace{-1.0em}
\caption{Ablation studies of the filter $\mathcal{F}(\cdot)$ in Phase 2 and Phase 3 on the Synthetic-I dataset.}
\label{tab:comp}
\vspace{-1.8 em}
\end{table}
%%%%%%%%%%%%%%%%%%%%%%%%%  Table4 Loss
\begin{table}[ht]
\begin{center}
	% \resizebox{0.45\textwidth}{!}{
		\resizebox{1.0\linewidth}{!}{
			\begin{tabular}{llll|llll}
				\toprule
				$\mathcal{L}_{mag}$ &$\mathcal{L}_{dr}$ & $\mathcal{L}_{edge}$ & $\mathcal{L}_{Sobel}$ & RMSE$\downarrow$   & PSNR$\uparrow$ & SSIM$\uparrow$  & LPIPS$\downarrow$   \\ \midrule
				\Checkmark &\XSolidBrush & \XSolidBrush & \XSolidBrush & 0.0678 & 22.05 & 0.9317 & 0.1121  \\
				\Checkmark &\Checkmark & \XSolidBrush & \XSolidBrush & 0.0613 & 24.91 & 0.9405 & 0.0783 \\
				\Checkmark &\Checkmark & \XSolidBrush & \Checkmark & \underline{0.0606} & \underline{25.06} & \underline{0.9487} & \underline{0.0687} \\
				\Checkmark &\Checkmark & \Checkmark & \XSolidBrush & \textbf{0.0594} & \textbf{25.49} & \textbf{0.9536} & \textbf{0.0535}\\ \bottomrule
			\end{tabular}
		}
	\end{center}
	\vspace{-1.0em}
	\caption{Ablation studies of loss functions on the Synthetic-I dataset.}
	\label{tab:loss}
	\vspace{-2.0em}
\end{table}
%%%%%%%%%%%%%%%%%%%%%%%%%%%%%%%%%%%%%%
\subsection{Quantitative Comparisons} \label{sec:quan}
\subsubsection{Comparisons on Synthetic Datasets.} 
We compare our method with existing approaches and report the experimental results in Tab.~\ref{tab:sys}.
On the Synthetic-I dataset (``clean scenario''), 
our method performs superior to the recent best method MDLMM~\cite{singh2023multi} on magnification error and visual quality, with RMSE, PSNR, SSIM, and LPIPS values of 0.0594 \vs 0.0651, 25.49 dB \vs 24.84 dB, 0.9536 \vs 0.9173, and 0.0535 \vs 0.1228, respectively. On Synthetic-II and Synthetic-III (``noise and blur scenes''),
EulerMormer still shows significant performance gains in the frames with Poisson noise and Gaussian blur.

\subsubsection{Comparisons on Real-World Datasets.}
From Tab.~\ref{tab:maniqa}, previous works with traditional narrowband filters~\cite{zhang2017video,takeda2018jerk,takeda2019video} have lower MANIQA scores than ours. The MANIQA metric~\cite{yang2022maniqa} mainly evaluates visual distortion levels.
For example, compared to the previous best method Anisotropy~\cite{takeda2019video}, we achieve 0.6920 \vs 0.6872, 0.6760 \vs 0.6634, and 0.7316 \vs 0.7288 on Static, Dynamic, and Fabric datasets, respectively.

%%%%%%%%%%%%%%%%%%%%%%%%%%%%%%%%%%%%%%%%%%%%%
\begin{figure*}[t!]
	\begin{center}
		\includegraphics[width=1\linewidth]{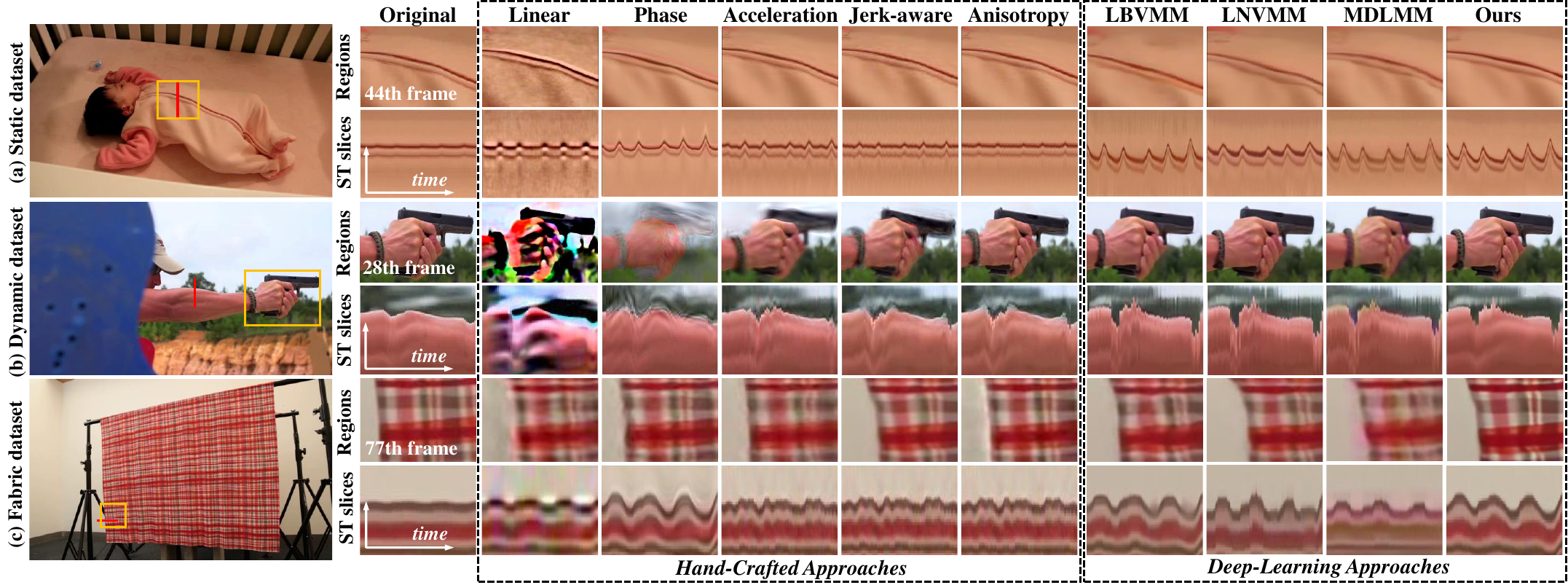}
		\vspace{-2em}
		\caption{Qualitative results of our method with existing methods on (a) Static, (b) Dynamic and (c) Fabric datasets with magnification factors $\alpha$ of 20, 10, and 20, respectively. We highlight spatial regions where motion occurs and provide spatiotemporal (ST) slices of magnified motion for better comparison.
		} 
		\label{fig:real}
	\end{center}
	\vspace{-1.7 em}
\end{figure*}
%%%%%%%%%%%%%%%%%%%%%%%%%%%%%%%%%%%%%%%%  
\begin{figure*}[t!]
	\begin{center}
		\includegraphics[width=1\linewidth]{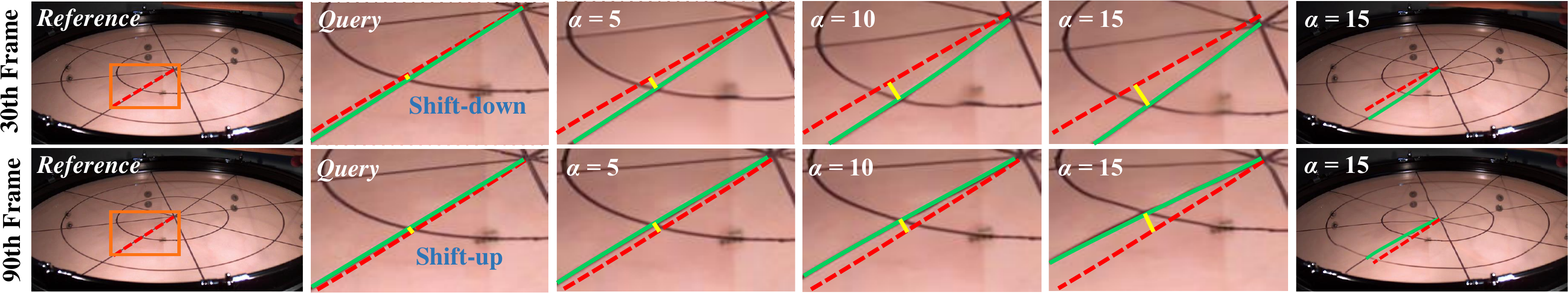}
		\vspace{-2 em}
		\caption{Magnification visualization of the \textit{drum} video from the Static dataset.
			We randomly sample two frames with shift-down and shift-up motion (the 30th and 90th frames in the video).
			Our method achieves reliable video motion magnification under different magnification factors $\alpha$.} 
		\label{fig:mag}
	\end{center}
	\vspace{-1.8 em}
\end{figure*}
%%%%%%%%%%%%%%%%%%%%%%%%%%%%%%%%%%
\begin{figure*}[t!]
	\begin{center}
		\includegraphics[width=1\linewidth]{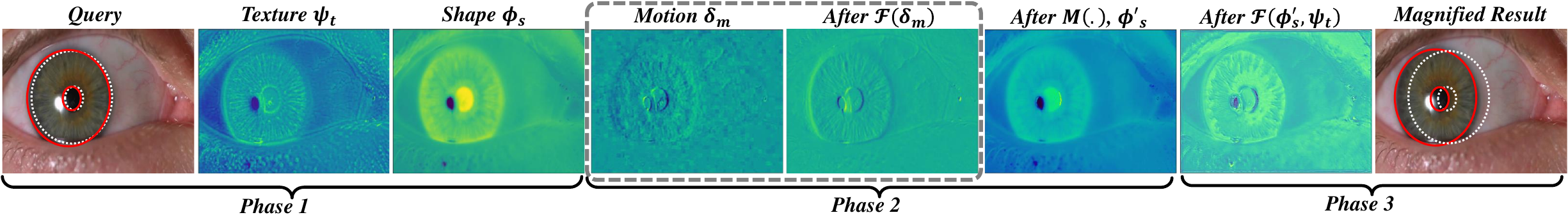}
		\vspace{-2 em}
		\caption{Dataflow of our method pipeline with the \textit{eye} video from the Dynamic dataset. The disentangled texture, shape, and motion feature maps have distinguishable vision characteristics. The dynamic filter $\mathcal{F}(\cdot)$ effectively erases the noises in the static-dynamic field of the image and refines the texture-shape joint refinement process.}
		\label{fig:vis}
	\end{center}
	\vspace{-2 em}
\end{figure*}
%%%%%%%%%%%%%%%%%%%%%%%%%%%%%%%%%%%%%%%%%%%%%
\vspace{-0.1 em}
\subsection{Ablation Studies} \label{sec:ab}
\subsubsection{Effectiveness of Filter $\mathcal{F}(\cdot)$.}
We test the dynamic filter $\mathcal{F}(\cdot)$ in Phase 2 and Phase 3 separately.
Observing Tab.~\ref{tab:comp}, removing $\mathcal{F}(\cdot)$ from the model significantly decreases both accuracy and quality of the magnification (\eg, removing $\mathcal{F}(\cdot)$ decreases the perceptual quality LPIPS from 0.0535 to 0.1170 in Phase 2 and increases the magnification error RMSE from 0.0594 to 0.0747 in Phase 3).
Moreover, we deeply discuss the two core components of $\mathcal{F}(\cdot)$, DMF and MFR. A more comprehensive analysis highlights the significant roles played by the DMF and MGR modules in denoising and artifacts-freeing (\eg, DMF improves PSNR from 24.72 to 25.49 in Phase 2, and MGR improves SSIM from 0.9437 to 0.9536 in Phases 3), thus validating the effectiveness of entire dynamic filter $\mathcal{F}(\cdot)$ in this task.

\subsubsection{Impact of Top-$k$ in Filter $\mathcal{F}(\cdot)$.}
To investigate the impact of the Top-$k$ operator in Filter $\mathcal{F}(\cdot)$, we test $k\in\{$1, 3, 5, 7, 9, 11$\}$. Here, $k \in [$0, $\hat{C}]$ and $\hat{C}$ = 12 in our experiment setup. 
From Fig.~\ref{fig:topk}, while $k$ = 1, it leads to significant sparsity of similarity-based attention matrix, resulting in a large error, \ie, RMSE of 0.0648. While $k$ = 11, there is a large error too, \ie, RMSE of 0.0621. Hence, an appropriate value of $k$ contributes to the balance of the attention sparsity calculation and magnification denoising.
As a result, we set $k$ = 7 with the lowest RMSE of 0.0594 as the optimal setting.

\subsubsection{Effect of Loss Function.}
Tab.~\ref{tab:loss} reports the ablation studies of different loss functions. 
Based on the basis $\mathcal{L}_{mag}$, the introduction of disentangled representation loss $\mathcal{L}_{dr}$ significantly improves the robustness of magnification, \ie, PSNR is improved from 22.05 dB to 24.91 dB. Moreover, applying the new edge detection loss $\mathcal{L}_{edge}$ yields gains of 0.58 dB and 0.0131 for PSNR and SSIM, respectively.
Comparing it with the well-known Sobel loss $\mathcal{L}_{Sobel}$~\cite{zheng2020image}, which focuses solely on horizontal and vertical edges, our $\mathcal{L}_{edge}$ incorporates the LoG operator for noise smoothing and edge detection, demonstrates better noise robustness, edge continuity, and effective extraction of low-contrast magnified motion boundaries (\eg, PSNR gain for $\mathcal{L}_{LoG}$ is 0.43 dB more than that for $\mathcal{L}_{Sobel}$).

\subsection{Qualitative Analysis}\label{sec:qual}
(1) \textbf{Magnification visualization comparisons.} As shown in Fig.~\ref{fig:real}, Linear~\cite{wu2012eulerian} and Phase~\cite{wadhwa2013phase} exhibit significant distortion and ringing artifacts; Acceleration~\cite{zhang2017video}, Jerk-aware~\cite{takeda2018jerk} and Anisotropy methods~\cite{takeda2019video} show insufficient amplification amplitude, and the other learning-based methods~\cite{oh2018learning,singh2023lightweight,singh2023multi} show flickering artifacts and motion distortion originating from their spatial inconsistency (CNN's local receptive field).
In contrast, we achieve more robust results, noticeably improving artifacts and distortions while achieving satisfactory magnification amplitude.
(2) \textbf{Magnification factor $\alpha$.} Fig.~\ref{fig:mag} illustrates the magnified results of shift-up and shift-down setups of the \emph{drum} surface. EulerMormer achieves reliable magnification at different magnification levels. 
(3) \textbf{Magnification dataflow.}
Fig.~\ref{fig:vis} displays the dataflow of EulerMormer. The disentangled texture, shape, and motion feature maps have distinguishable vision characteristics.
Please pay attention to the dynamic filter $\mathcal{F}(\cdot)$ in Phase 2, which effectively eliminates noise in the static field of the motion $\delta_m$ while preserving important motion information in the dynamic field.
On this basis, the visualization of $\mathcal{F}(\phi'_s(x,t),\psi_t(x,t))$ also validates the ability of $\mathcal{F}(\cdot)$ to process the texture-magnified shape joint refinement.

\section{Conclusion}\label{sec:conclusion}
In this paper, we have introduced EulerMormer, a novel Transformer-based end-to-end framework designed for video motion magnification tasks from the Eulerian perspective, aiming to provide more robust magnification effects.
The core of EulerMormer lies in embedding a dedicated dynamic filter within Transformer, enabling static-dynamic field adaptive denoising for motion representation and recoupling representation refinement.
To validate the model's accuracy and robustness, we collect a synthetic dataset with magnification factors, Poisson noise, and Gaussian blur to provide comprehensive quantitative evaluations. Extensive quantitative and qualitative experiments demonstrate that EulerMormer outperforms state-of-the-art approaches.

\subsubsection{Acknowledgments.}
This work was supported by the National Key R\&D Program of China (2022YFB4500600), the National Natural Science Foundation of China (62272144, 72188101, 62020106007, and U20A20183), and the Major Project of Anhui Province (202203a05020011).

\bibliography{aaai24}

\end{document}